\documentclass{article}
\usepackage{ijcai20}

\usepackage{times}

\usepackage{soul}
\usepackage{url}
\usepackage[hidelinks]{hyperref}
\usepackage[utf8]{inputenc}
\usepackage[small]{caption}
\usepackage{graphicx}
\usepackage{amsmath}
\usepackage{booktabs}
\usepackage{algorithm}
\usepackage{amsfonts}

\usepackage{algpseudocode}
\usepackage{latexsym}
\usepackage{multirow}
\usepackage{varioref}
\usepackage{xspace}
\usepackage{bm}

\usepackage{makecell}
\newcommand{\eg}{\emph{e.g.,}\xspace}
\newcommand{\ie}{\emph{i.e.,}\xspace}
\newcommand{\etc}{\emph{etc.}\xspace}
\newcommand{\baby}{PML$^3$\textsc{er}\xspace}

\title{Recovering Accurate Labeling Information from Partially Valid Data for Effective Multi-Label Learning}

\author{
Ximing Li$^{1,2}$ \And 
Yang Wang$^{3,4,}$\footnote{Yang Wang is the Corresponding Author, denoted by *}
\affiliations
$^1$ College of Computer Science and Technology, Jilin University, China\\
$^2$ Key Laboratory of Symbolic Computation and Knowledge Engineering of Ministry of Education, Jilin University, China\\
$^3$ Key Laboratory of Knowledge Engineering with Big Data, Ministry of Education, Hefei University of Technology, China\\
$^4$ School of Computer Sci \& Information Engineering, Hefei University of Technology, China\\
\emails
liximing86@gmail.com,
yangwang@hfut.edu.cn
}

\begin{document}

\maketitle

\begin{abstract}
Partial Multi-label Learning (PML) aims to induce the multi-label predictor from datasets with noisy supervision, where each training instance is associated with several candidate labels but only partially valid. To address the noisy issue, the existing PML methods basically recover the ground-truth labels by leveraging the ground-truth confidence of the candidate label, \ie the likelihood of a candidate label being a ground-truth one. However, they neglect the information from non-candidate labels, which potentially contributes to the ground-truth label recovery. In this paper, we propose to recover the ground-truth labels, \ie estimating the ground-truth confidences, from the label enrichment, composed of the relevance degrees of candidate labels and irrelevance degrees of non-candidate labels. Upon this observation, we further develop a novel two-stage PML method, namely \emph{\underline{P}artial \underline{M}ulti-\underline{L}abel \underline{L}earning with \underline{L}abel \underline{E}nrichment-\underline{R}ecovery} (\baby), where in the first stage, it estimates the label enrichment with unconstrained label propagation, then jointly learns the ground-truth confidence and multi-label predictor given the label enrichment. Experimental results validate that \baby outperforms the state-of-the-art PML methods.
\end{abstract}

\section{Introduction}
\label{1}

\begin{figure*}[t]
\includegraphics[width=0.97\textwidth]{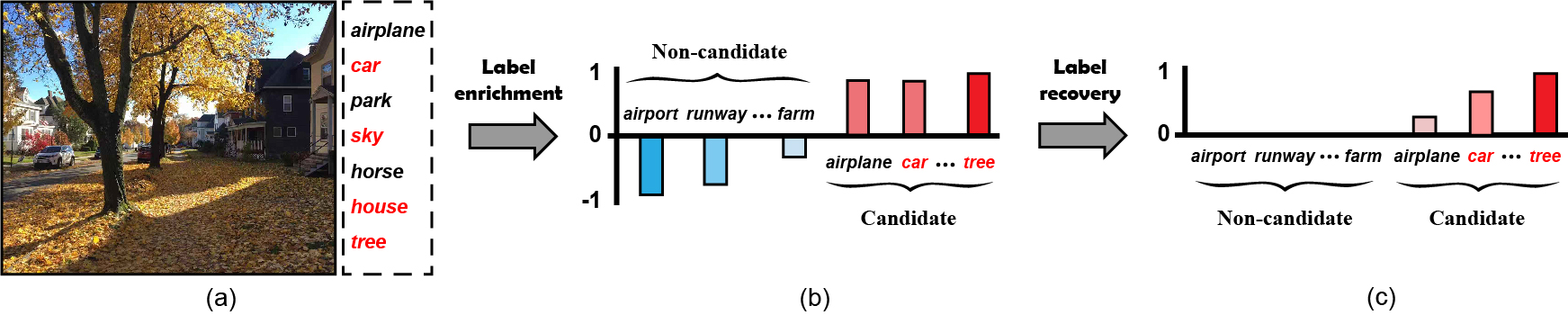}
\centering
\caption{The basic idea of \baby. (a) An example of PML image instance annotated with 7 candidate labels, with only 4 of them are ground-truth labels, \ie the ones in red (best view in color). (b) indicates the label enrichment,  \ie estimating both the relevance degrees  of the candidate labels within (0,1) range, and irrelevance degrees of the non-candidate labels within (-1,0) range. (c) indicates the label recovery, \ie estimating the ground-truth confidences of candidate labels from the label enrichment. The candidate label ``airplane'' is more likely to be an irrelevant noisy label, since its highly correlated labels, \eg ``airport'' and ``runway'', are with higher irrelevance degrees, to filter out the noisy candidate label.}
\label{ExamplePML}
\end{figure*}

\textbf{P}artial \textbf{M}ulti-label \textbf{L}earning (\textbf{PML}), a novel learning paradigm with noisy supervision, draws increasing attention from the machine learning community \cite{PML2019,Juan2019}. It refers to induce the multi-label predictor from PML datasets, with each training instance associated with multiple candidate labels that are only partially valid. The PML datasets are available in many real-world applications, where collecting the accurate supervision is quite expensive for many scenarios, \eg crowdsourcing annotations. To visualize this, we illustrate a PML image instance in Fig.\ref{ExamplePML}(a): an annotator may roughly select more candidate labels, so as to cover all ground-truth labels but ineluctably with several irrelevant ones, imposing big challenges for learning with such noisy PML training instances.

Formally speaking, given a PML dataset ${\mathcal{D}} = \{\left(\mathbf{x}_i,\mathbf{y}_i\right)\}_{i=1}^{i=n}$ with  $n$ instances and $l$ labels, where $\mathbf{x}_i \in \mathbb{R}^d$ denotes the feature vector and $\mathbf{y}_i \in \{0,1\}^l$ the candidate label set of $\mathbf{x}_i$. For $\mathbf{y}_i$, the value of 0 or 1 represents the corresponding label to be a non-candidate or candidate label. Let $\mathbf{y}^* \in \{0,1\}^l$ denote the (unknown) ground-truth label sets of instances. Specifically, for each instance $\mathbf{x}_i$, its corresponding $\mathbf{y}^*_i$ is covered by the candidate label set $\mathbf{y}_i$, \ie $\mathbf{y}^*_i \subseteq \mathbf{y}_i$. Accordingly, the task of PML aims to induce a multi-label predictor $f(\mathbf{x}): \mathbb{R}^d \to \{0,1\}^l$ from $\mathcal{D}$.

To solve the problem, several typical attempts have been made \cite{PML2018,Yu2018,PML2019,Juan2019}, where the basic idea is to recover the ground-truth labels by leveraging the \emph{ground-truth confidence} of the candidate label, \ie the likelihood of a candidate label being a ground-truth one, and learning with it, instead of the candidate label. For example, an early framework of PML \cite{PML2019} estimates the ground-truth confidences via label propagation over an instance neighbor graph, following the intuition that neighboring instances tend to share the same labels; another work \cite{Juan2019} recovered the ground-truth confidences by decomposing the candidate labels under a low-rank scheme. 

Revisiting the existing PML methods, we find that they basically estimate the ground-truth confidence around candidate label annotations, but neglect the information from non-candidate labels, which potentially contributes to the ground-truth label recovery. The irrelevance degree of the non-candidate label, \ie the degree of a non-candidate label being irrelevant with the instance, is contributive to distinguish the irrelevant noisy labels within candidate label sets, since any candidate label tends to be an irrelevant noisy one if its highly correlated labels are given higher irrelevance degrees. For example, in Fig.\ref{ExamplePML}(b) and (c), the candidate label ``airplane'' is more likely to be an irrelevant noisy label, since its highly correlated labels, \eg ``airport'' and ``runway'', are with higher irrelevance degrees for the example image instance.

Based on the above analysis, we propose to estimate the ground-truth confidences over both candidate and non-candidate labels. In particular, we develop a novel two-stage PML method, namely \textbf{P}artial \textbf{M}ulti-\textbf{L}abel \textbf{L}earning with \textbf{L}abel \textbf{E}nrichment-\textbf{R}ecovery (\textbf{\baby}). On the first stage, we estimate the \emph{label enrichment}, composed of the relevance degrees of candidate labels (\ie the complementary definition of irrelevance degree) and irrelevance degrees of non-candidate ones, using an unconstrained label propagation procedure. In the second stage, we jointly train the ground-truth confidence and multi-label predictor given the label enrichment learned from the first stage. We conduct extensive experiments to validate the effectiveness of \baby. 

The contributions of this paper are summarized below:
\begin{itemize}
    \item We propose \baby by leveraging both the information from candidate and non-candidate labels.  
    \item The \baby estimates the label enrichment using unconstrained label propagation, and then trains the multi-label predictor with label recovery simultaneously.
    \item We conduct extensive experiments to validate the effectiveness of \baby. 
\end{itemize}

\section{Related Work}
\label{2}

\subsection{Partial Multi-label Learning}
Abundant researches towards Partial Label Learning (PLL) have been made, where each training instance is annotated with a candidate label set but only one is valid \cite{Cour2011,Liu2012,Chen2014,Zhang2017,Yu2017,Wu2018,Gong2018,Chen2018,Feng2018,Feng2019,Feng2019_2,KDD2019}. The core idea of PLL follows the spirit of disambiguation, \ie identifying the ground-truth label from the candidate label set for each instance. In some cases, PLL can be deemed as a special case of PML, where the ground-truth label number is fixed to one. Naturally, PML is more challenging than PLL, even the number of ground-truth labels is unknown.

The existing PML methods mainly recover the ground-truth labels by estimating the ground-truth confidences  \cite{PML2018,Yu2018,PML2019,Juan2019}. Two PML methods are proposed \cite{PML2018}, \ie Partial Multi-label Learning with \emph{l}abel \emph{c}orrelation (PML-\emph{lc}) and \emph{f}eature \emph{p}rototype (PML-\emph{fp}), which are upon the ranking loss objective weighted by ground-truth confidences. The other one \cite{Juan2019}, namely Partial Multi-label Learning by Low-Rank and Sparse decomposition (PML-LRS), trains the predictor with ground-truth confidences under the low-rank assumption. Besides them, the two-stage PML framework \cite{PML2019}, \ie PARTIal multi-label learning via Credible Label Elicitation (P\textsc{article}), estimates the ground-truth confidences by label propagation, and then trains the predictor over candidate labels with high confidences only. Two traditional methods of victual Label Splitting (VLS) and Maximum A Posteriori (MAP) are used in its second stage, leading to two versions, \ie P\textsc{article}-V\textsc{ls} and P\textsc{article}-M\textsc{ap}.

Orthogonal to those methods, our \baby estimates the ground-truth confidence using the label enrichment involving both candidate and non-candidate labels to recover more accurate supervision.

\subsection{Learning with Label Enrichment}
Learning with label enrichment, also known as label enhancement, explores richer label information, \eg the label membership degrees of instances, to enhance the performance \cite{LE2006_1,LE2006_2,LE2015,LE2016,LE2018}. The existing methods achieve the label enrichment using the similarity knowledge of instances by various schemes, such as fuzzy clustering \cite{LE2006_1}, label propagation \cite{LE2015}, manifold learning \cite{LE2016} \etc They have been applied to the paradigms of multi-label learning and label distribution learning with accurate annotations.

Our \baby also refers to the label enrichment, but it works on the scenario of PML under noisy supervision.

\section{The \baby Algorithm}
\label{3}

Following the notations in Section 1, we denote $\mathbf{X}=[\mathbf{x}_1,\cdots,\mathbf{x}_n]^{\top} \in \mathbb{R}^{n \times d}$ and $\mathbf{Y}=[\mathbf{y}_1,\cdots,\mathbf{y}_n]^{\top} \in \{0,1\}^{n \times l}$ the instance matrix and candidate label matrix, respectively. For each instance $\mathbf{x}_i$, $y_{ij}=1$ means that the $j$-th label is a candidate label, otherwise it is a non-candidate one.

Given a PML dataset $\mathcal{D} = \{\mathbf{X,Y}\}$, our \baby first estimates the label enrichment, which describes the relevance degrees of candidate labels and irrelevance degrees of non-candidate ones. Specifically, for each instance $\mathbf{x}_i$, we denote the corresponding label enrichment $\mathbf{\widehat y}_i = [\widehat y_{i1},\cdots,\widehat y_{il}]^{\top}$ as follows:
\begin{equation}
\label{Eq3-1}
\widehat y_{ij} \in
\left\{
             \begin{array}{lr}
             [0,1]\:,   \:\:\:\:\:\:\:\: \mathbf{if} \:\: y_{ij} =1 \\\relax
             [ - 1,0]\:, \:\:\:\:  \mathbf{otherwise}
             \end{array}
\right.
\quad \quad \forall j \in [l],
\end{equation}
referring to the example image instance in Fig.\ref{ExamplePML}(b). Further, we denote by $\mathbf{\widehat Y}=[\mathbf{\widehat y}_1,\cdots,\mathbf{\widehat y}_n]^{\top}$ the label enrichment matrix. Then, \baby induces the multi-label predictor from the enrichment version of PML dataset $\mathcal{\widehat D} = \{\mathbf{X,\widehat Y}\}$.

We now describe the two stages of \baby, \ie \emph{label enrichment by unconstrained propagation} and \emph{jointly learning the ground-truth confidence and multi-label predictor}.

\subsection{Label Enrichment by Unconstrained Propagation}
\label{3.1}

In the first stage, \baby estimates the label enrichment matrix $\mathbf{\widehat Y}$. Inspired by \cite{LE2015,PML2019}, we develop an unconstrained label propagation procedure, which estimates the labeling degrees by progressively propagating annotation information over a weighted \emph{k}-Nearest Neighbor (\emph{k}NN) graph over the instances. The intuition is that the candidate and non-candidate labels with more accumulations during the \emph{k}NN propagation tend to be with higher relevance degrees and lower irrelevance degrees, respectively.

We describe the detailed steps of the unconstrained label propagation procedure below.

\vspace{2pt}
\noindent\textbf{[Step 1]}: After constructing the \emph{k}NN graph $\mathbf{\Omega}$, for each instance $\mathbf{x}_i$, we compute its reconstruction weight vector of \emph{k}NNs, \ie $\mathbf{v}_i=[v_{i1},\cdots,v_{in}]^{\top}\in\mathbb{R}^n$, by minimizing the following objective:
\begin{eqnarray}
\!\!\!\!\!\!\!\!\!\!\!\!\!\!\!\!\!\!\!\!\! && \mathop{\mathbf{min}} \limits_{\mathbf{v}_i} \| \mathbf{x}_i - \mathbf{X}^{\top}\mathbf{v}_i \|_2^2 \nonumber \\
\!\!\!\!\!\!\!\!\!\!\!\!\!\!\!\!\!\!\!\!\! && \: {\rm{\mathbf{s.t.}}} \:\: v_{ij} \geq 0 \quad \forall j \in \mathbf{\Omega}(\mathbf{x}_i), \:\: v_{ij} = 0 \quad \forall j \not \in \mathbf{\Omega}(\mathbf{x}_i),
\label{Eq3-1-1}
\end{eqnarray}
where $\mathbf{\Omega}(\mathbf{x}_i)$ denotes the \emph{k}NNs of $\mathbf{x}_i$. This objective can be efficiently solved by any off-the-shelf Quadratic Programming (QP) solver\footnote{We apply the public QP solver of \emph{mosek} downloaded at {https://www.mosek.com/}}. Repeat to solve the problem of Eq.\eqref{Eq3-1-1} for each instance, we can achieve the reconstruction weight matrix $\mathbf{V} = [\mathbf{v}_1,\cdots,\mathbf{v}_n]^{\top}\in\mathbb{R}^{n \times n}$. Then, we normalize $\mathbf{V}$ by row, \ie $\mathbf{V} \leftarrow \mathbf{VD}^{-1}$, where $\mathbf{D} = {\rm{diag}}[d_1,\cdots,d_n], \: d_i = \sum_{j=1}^n v_{ij}$.

\vspace{2pt}
\noindent \textbf{[Step 2]}: Following the idea that the relationships of instances can be translated to their associated labels, we can enrich the labeling information by propagating over $\mathbf{\Omega}$ with $\mathbf{V}$. Formally, we denote by $\mathbf{F} = [\mathbf{f}_1,\cdots,\mathbf{f}_n]^{\top} \in \mathbb{R}_+^{n \times l}$ the label propagation solution, which is initialized by the candidate label matrix $\mathbf{Y}$, \ie $\mathbf{F}^{(0)} = \mathbf{Y}$. Then, $\mathbf{F}$ can be iteratively updated by propagating over $\mathbf{\Omega}$ with $\mathbf{V}$ until convergence. At each iteration $t$, the update equation is given by:
\begin{equation}
\mathbf{F}^{(t)} = \alpha \cdot \mathbf{V}^{\top}\mathbf{F}^{(t-1)} + (1-\alpha) \cdot \mathbf{F}^{(0)},
\label{Eq3-1-2}
\end{equation}
where $\alpha \in [0,1]$ is the propagation rate. To avoid overestimate non-candidate labels, we normalize each row of $\mathbf{F}^{(t)}$:
\begin{eqnarray}
\!\!\!\!\!\!\!\!\!\!\!\!\!\!\!\!\!\!\!\!\! && f_{ij}^{(t)} = \mathbf{min}\left(1\:,\:\frac{f_{ij}^{(t)} - \mathbf{vmin}({\mathbf{f}_i^{(t)}})} {\mathbf{cvmax}(\mathbf{f}_i^{(t)}) - \mathbf{vmin}(\mathbf{f}_i^{(t)})} \right), \nonumber \\
\!\!\!\!\!\!\!\!\!\!\!\!\!\!\!\!\!\!\!\!\! && \quad \quad \quad \quad \quad \quad \quad \quad \quad \quad \quad \quad \quad \quad  \forall i \in [n], \: \forall j \in [l],
\label{Eq3-1-3}
\end{eqnarray}
where $\mathbf{vmin}(\mathbf{f}_i^{(t)})$ denotes the minimum of $\mathbf{f}_i^{(t)}$, and $\mathbf{cvmax}(\mathbf{f}_i^{(t)})$ the maximum over candidate labels in $\mathbf{f}_i^{(t)}$.

\vspace{2pt}
\noindent \textbf{[Step 3]}: After obtaining the optimal $\mathbf{F}$, denoted by $\mathbf{F}^* = [f^*_{ij}]_{n \times l}$, we compute the label enrichment matrix $\mathbf{\widehat Y}$ as follows:
\begin{equation}
\widehat{y}_{ij} =
\left \{
    \begin{array}{lr}
    f_{ij}^* \:, \:\:\:\:\:\:\:\:\:\:\:\:  \mathbf{if} \:\: y_{ij}=1 \\\relax
    f_{ij}^*-1 \:, \:\:\:\: \mathbf{otherwise}
    \end{array}
\right.
\quad \forall i \in [n].\:\forall j \in [l]
\label{Eq3-1-4}
\end{equation}
For clarity, we summarize the unconstrained label propagation for $\mathbf{\widehat Y}$ in \emph{Algorithm \ref{Alg1}}.

\begin{algorithm}[t]
  \caption{Unconstrained label propagation for $\mathbf{\widehat Y}$}
  \label{Alg1}
  \begin{algorithmic}[1]
    \Require{The PML dataset ${\mathcal{D}} = \{\mathbf{X,Y}\}$, nearest neighbor number $k$ and propagation rate $\alpha$;}
    \Ensure{The label enrichment matrix $\mathbf{\widehat Y}$.}
    \State{}Construct the \emph{k}NN graph of $\mathbf{X}$, and initialize $\mathbf{F}^{(0)}$ by $\mathbf{Y}$
    \State{}Compute $\mathbf{V}$ by solving Eq.\eqref{Eq3-1-1} for each instance, and then normalize it by row
    \State{}\textbf{While not convergence Do}
    \State{}$\quad$ Update $\mathbf{F}$ using Eq.\eqref{Eq3-1-2}
    \State{}$\quad$ Normalize $\mathbf{F}$ using Eq.\eqref{Eq3-1-3}
    \State{}\textbf{End While}
    \State{}Compute $\mathbf{\widehat Y}$ using Eq.\eqref{Eq3-1-4}
  \end{algorithmic}
\end{algorithm}

\begin{algorithm}[t]
  \caption{Predictive model induction for \baby}
  \label{Alg2}
  \begin{algorithmic}[1]
    \Require{The enriched PML dataset ${\mathcal{\widehat D}} = \{\mathbf{X,\widehat Y}\}$, regularization parameters $\{\lambda_1,\lambda_2\}$;}
    \Ensure{The predictive parameter matrix $\mathbf{W}$.}
    \State{}Initialize $\{\mathbf{C,B,W},\mathbf{\widehat B,\Theta}\}$
    \State{}\textbf{While not convergence Do}
    \State{}$\quad$ Update $\mathbf{C}$ using Eq.\eqref{Eq3-2-6}
    \State{}$\quad$ \textbf{For $t=1$} to $N_{iter}$
    \State{}$\quad$ $\quad$ Update $\mathbf{\widehat B,B,\Theta}$ using Eqs.\eqref{Eq3-2-12}, \eqref{Eq3-2-13} and \eqref{Eq3-2-14}
    \State{}$\quad$ \textbf{End For}
    \State{}$\quad$ Update $\mathbf{W}$ using Eq.\eqref{Eq3-2-16}
    \State{}\textbf{End While}
  \end{algorithmic}
\end{algorithm}

\subsection{Jointly Learning the Ground-truth Confidence and Multi-label Predictor}
\label{3.2}

In the second stage, \baby jointly trains the ground-truth confidence matrix $\mathbf{C} = [c_{ij}]_{n \times l} \in [0,1]^{n \times l}$ and multi-label predictor given the enrichment version of PML dataset $\mathcal{\widehat D} = \{\mathbf{X,\widehat Y}\}$.

First, we aim to recover $\mathbf{C}$ from $\mathbf{\widehat Y}$ by leveraging the following minimization:
\begin{eqnarray}
\!\!\!\!\!\!\! && \mathop{\mathbf{min}} \limits_{\mathbf{C,B}} \| \mathbf{\widehat{Y}}-\mathbf{CB} \|_F^2 + \lambda_1 \|\mathbf{B}\|_*\nonumber \\
\!\!\!\!\!\!\! && \:\:\:\: {\rm{\textbf{s.t.}}} \:\:\:\: \mathbf{0}_{n \times l} \preceq \mathbf{C} \preceq \mathbf{Y},
\label{Eq3-2-1}
\end{eqnarray}
where $\mathbf{B} = [b_{ij}]_{l \times l}$ denotes the label correlation matrix, and $\mathbf{0}_{n \times l}$ the all zero matrix. Specifically for capturing local label correlations, we utilize the nuclear regularizer for $\mathbf{B}$, \ie $\|\mathbf{B}\|_*$, with the regularization parameter $\lambda_1$.

Second, we aim to train a linear multi-label predictor with $\mathbf{C}$ by leveraging a least square minimization with a squared Frobenius norm regularization:
\begin{equation}
\mathop{\mathbf{min}} \limits_{\mathbf{W}} \| \mathbf{C-XW} \|_F^2 + \lambda_2 \| \mathbf{W} \|_F^2
\label{Eq3-2-3}
\end{equation}
where $\mathbf{W} \in \mathbb{R}^{d \times l}$ is the predictive parameter matrix, and $\lambda_2$ the regularization parameter.

By combining Eqs.\eqref{Eq3-2-1} and \eqref{Eq3-2-3}, we achieve the overall objective as follows:
\begin{eqnarray}
\!\!\!\!\!\!\!\!\!\!\!\!\!\! && \mathop{\mathbf{min}} \limits_{\mathbf{C,B,W}} \| \mathbf{\widehat{Y}}-\mathbf{CB} \|_F^2 + \| \mathbf{C-XW} \|_F^2 + \lambda_1 \|\mathbf{B}\|_* + \lambda_2 \| \mathbf{W} \|_F^2 \nonumber \\
\!\!\!\!\!\!\!\!\!\!\!\!\!\! && \:\:\:\: {\rm{\textbf{s.t.}}} \:\:\:\: \mathbf{0}_{n \times l} \preceq \mathbf{C} \preceq \mathbf{Y}
\label{Eq3-2-4}
\end{eqnarray}

\paragraph{Discussion on Recovering $\mathbf{C}$ from $\mathbf{\widehat Y}$.}
Referring to Eq.\eqref{Eq3-2-1}, we jointly learn the ground-truth confidence matrix $\mathbf{C}$ and label correlation matrix $\mathbf{B}$ by minimizing their reconstruction error of $\mathbf{\widehat Y}$. By omitting the regularizers, it can be roughly re-expressed below:
\begin{eqnarray}
\!\!\!\!\!\!\!\!\!\!\!\!\!\!\!\!\!\!\!\!\! && \mathop{\mathbf{min}} \limits_{\mathbf{C,B}}  \sum \nolimits_{i=1} ^{n} \sum \nolimits_{j=1} ^l \left( \widehat y_{ij} - \sum \nolimits_{h=1}^l c_{ih}b_{hj} \right)^2 \nonumber \\
\!\!\!\!\!\!\!\!\!\!\!\!\!\!\!\!\!\!\!\!\! && \:\: {\rm{\textbf{s.t.}}} \:\:\:\:  c_{ij} \in [0,1], \:\:  \forall y_{ij}=1;  \quad  c_{ij} =0, \:\: \forall y_{ij}=0 \nonumber
\end{eqnarray}
Obviously, for each component $\widehat y_{ij}$, it contributes that any $c_{ih}$ tends to be larger or smaller given larger value of $b_{hj}$ (\ie  a higher correlation between label $j$ and $h$), if $\widehat y_{ij}$ corresponds to a candidate label ($\geq 0$) or a non-candidate one ($\leq 0$). That is, we actually recover $\mathbf{C}$ using the information from candidate and non-candidate labels simultaneously.

\subsubsection{Optimization}
\label{3.2.1}

As directly solving the objective of Eq.\eqref{Eq3-2-4} is intractable, we optimize the variables of interest, \ie $\{\mathbf{C,B,W}\}$, via alternating fashion, by optimizing one variable with the other two fixed. Repeat this process until convergence or reaching the maximum number of iterations. We describe the update equations of $\{\mathbf{C,B,W}\}$ one by one.

\vspace{8pt}
\noindent{[\textbf{Update ${\mathbf{C}}$}]} When \{${\rm{\mathbf{B,W}}}$\} are fixed, the sub-objective with respect to ${\mathbf{C}}$ can be reformulated as follows:
\begin{align}
    & \mathop{\mathbf{min}} \limits_{\mathbf{C}} \| \mathbf{\widehat{Y}}-\mathbf{CB} \|_F^2 + \| \mathbf{C-XW} \|_F^2 \nonumber \\
    & \:\:\: {\rm{\textbf{s.t.}}} \:\:\:\: \mathbf{0}_{n \times l} \preceq \mathbf{C} \preceq \mathbf{Y}
\label{Eq3-2-5}
\end{align}
This optimization refers to a convex optimization, so as to achieve the following truncated update equation:
\begin{equation}
c_{ij} \leftarrow
\left \{
             \begin{array}{lr}
             0 \:, \:\:\:\:\:\:\:\:\:\:\:\: \mathbf{if} \:\: c'_{ij} \leq 0 \\\relax
             1 \:, \:\:\:\:\:\:\:\:\:\:\:\: \mathbf{if} \:\: c'_{ij} \geq 1\\\relax
             c'_{ij}\:, \:\:\:\:\:\:\:\:\: \mathbf{otherwise}
             \end{array}
        \right.
\quad \forall i \in [n],\: \forall j \in [l],
\label{Eq3-2-6}
\end{equation}
where $\mathbf{C}' = [c'_{ij}]_{n \times l}$:
\begin{equation*}
\mathbf{C'} = (\mathbf{\widehat{Y}B^{\top}} + \mathbf{XW})(\mathbf{BB^{\top}} + \mathbf{I}_l)^{-1}
\label{3-2-7}
\end{equation*}

\vspace{8pt}
\noindent{[\textbf{Update ${\mathbf{B}}$}]} When \{${\rm{\mathbf{C,W}}}$\} are fixed, the sub-objective with respect to ${\mathbf{B}}$ can be reformulated as follows:
\begin{equation}
\mathop{\mathbf{min}} \limits_{\mathbf{B}} \| \mathbf{\widehat{Y}-CB} \|_F^2 + \lambda_1 \| \mathbf{B} \|_*
\label{Eq3-2-10}
\end{equation}
Following the spirit of Alternating Direction Method of Multipliers (ADMM) \cite{ADMM2011}, we convert Eq.\eqref{Eq3-2-10} into an augmented Lagrange problem with an auxiliary matrix $\mathbf{\widehat B} \in \mathbb{R}^{l \times l}$ and Lagrange parameter matrix $\mathbf{\Theta} \in \mathbb{R}^{l \times l}$:
\begin{equation}
\mathop{\mathbf{min}} \limits_{\mathbf{\widehat B, B,\Theta}} \| \mathbf{\widehat{Y}-C\widehat{B}} \|_F^2 + \lambda_1 \| \mathbf{B} \|_* + \frac{\tau}{2} \| \mathbf{B-\widehat{B}} + \frac{\mathbf{\Theta}}{\tau} \|_F^2,
\label{Eq3-2-11}
\end{equation}
where $\tau$ is the penalty parameter. We perform an inner iteration to alternatively optimizing each of $\{\mathbf{\widehat B,B,\Theta}\}$ with the other two fixed. After some simple algebra, the update equations are formulated as follows:
\begin{equation}
\mathbf{\widehat B} \leftarrow (2\mathbf{C}^{\top}\mathbf{C} + \tau\mathbf{I}_l)^{-1}(2\mathbf{C}^{\top}\mathbf{\widehat{Y}} + \tau\mathbf{B} + \mathbf{\Theta})
\label{Eq3-2-12}
\end{equation}
For $\mathbf{B}$, it can be directly solved by
\begin{equation}
\mathbf{B} \leftarrow \mathbf{SVD}_{\frac{\lambda_1}{\tau}}\left(\mathbf{\widehat B} - \frac{\mathbf{\Theta}}{\tau}\right),
\label{Eq3-2-13}
\end{equation}
where $\mathbf{SVD}_{\frac{\lambda_1}{\tau}}(\cdot)$ is the singular thresholding with $\frac{\lambda_1}{\tau}$ \cite{SVD2010}. Then, for $\mathbf{\Theta}$, it can be updated by:
\begin{equation}
\mathbf{\Theta} \leftarrow \mathbf{\Theta} + \tau(\mathbf{B-\widehat{B}})
\label{Eq3-2-14}
\end{equation}

\vspace{8pt}
\noindent{[\textbf{Update ${\mathbf{W}}$}]} When \{${\rm{\mathbf{C,B}}}$\} are fixed, the sub-objective with respect to ${\mathbf{W}}$ can be reformulated as follows:
\begin{equation}
\mathop{\mathbf{min}} \limits_{\mathbf{W}} \| \mathbf{C-XW} \|_F^2 + \lambda_2 \| \mathbf{W} \|_F^2
\label{Eq3-2-15}
\end{equation}
The problem has an analytic solution as follows:
\begin{equation}
\mathbf{W} = (\mathbf{X}^{\top}\mathbf{X} + \lambda_2\mathbf{I}_d)^{-1}\mathbf{X}^{\top}\mathbf{C}
\label{Eq3-2-16}
\end{equation}

\begin{algorithm}[t]
  \caption{Full algorithm of \baby}
  \label{Alg3}
  \begin{algorithmic}[1]
    \Require{The PML dataset ${\mathcal{D}} = \{\mathbf{X,Y}\}$, $k=10$, $\alpha=0.5$ and $\{\lambda_1,\lambda_2\}$;}
    \Ensure{The predictive parameter matrix $\mathbf{W}$.}
    \State{}Compute the label enrichment matrix $\mathbf{\widehat Y}$ using \emph{Algorithm \ref{Alg1}}
    \State{}Optimize $\mathbf{W}$ using \emph{Algorithm \ref{Alg2}} given ${\mathcal{D}} = \{\mathbf{X,\widehat Y}\}$
  \end{algorithmic}
\end{algorithm}

For clarity, we summarize the procedure of this predictive model induction in \emph{Algorithm \ref{Alg2}}.

\subsection{\baby Summary}
\label{3.3}

We describe some implementation details of \baby. First, following \cite{PML2019}, we fix the parameters of the unconstrained label propagation as: $k=10$ and $\alpha=0.05$. Second, we empirically fix the penalty parameter $\tau$ of ADMM as 1. Third, the maximum iteration number of both ADMM loops is set to 5, as to be widely known, ADMM basically converges fast. Overall, the \baby algorithm is outlined in \emph{Algorithm \ref{Alg3}}.

\paragraph{Time Complexity.}
\label{3.3.1}

We also discuss the time complexity of \baby. First, in the unconstrained label propagation procedure, we require $\mathcal{O}(n^2d^2)$ time to construct the weighted \emph{k}NN graph, and $\mathcal{O}(T_1 n^2l)$ time to obtain the label enrichment matrix, referring to Eq.\eqref{Eq3-1-2}, where $T_1$ denotes its iteration number. Second, in predictive model induction, the major computational costs include matrix inversion and SVD, requiring roughly $\mathcal{O}(T_2(d^3+n^2l))$ time, where $T_2$ denotes the iteration number of the outer loop\footnote{Here, we omit the iteration numbers of inner ADMM loop for $\mathbf{B}$, since it converges very fast.}. Therefore, the total time complexity of \baby is $\mathcal{O}(n^2d^2+T_1 n^2l+T_2(d^3+n^2l))$.

\section{Experiment}
\label{4}

\subsection{Experimental Setup}
\label{4.1}

\begin{table}[t]
\small
\centering
\renewcommand\arraystretch{1.1}
\begin{tabular}{p{30pt}<{\centering}||p{24pt}<{\centering}p{24pt}<{\centering}p{20pt}<{\centering}p{26pt}<{\centering}p{32pt}<{\centering}}
\Xhline{1.5pt}
\multirow{1}{*}{\rm{\textbf{Dataset}}} & \multirow{1}{*}{$n$}  & \multirow{1}{*}{$d$}  & \multirow{1}{*}{$l$} & \multirow{1}{*}{\#{\rm{\textbf{AL}}}} & \multirow{1}{*}{\rm{\textbf{Domain}}} \\
\Xhline{0.5pt}
Genbase  & 662   & 1186   & 27   & 1.252    & biology  \\
Medical  & 978   & 1449   & 45   & 1.245    & text  \\
Arts  & 5000   & 462   & 26   & 1.636    & text \\
Corel5k  & 5000   & 499   & 374   & 3.522    & images  \\
Bibtex  & 7395   & 1836   & 159   & 2.406    & text  \\
\Xhline{1.5pt}
\end{tabular}
\caption{Statistics of the original multi-label datasets. ``\#AL'': average label number of each instance.}
\label{Datasets}
\end{table}

\begin{table*}[!ht]
\centering
\renewcommand\arraystretch{.92}
\scriptsize
\begin{tabular}{p{30pt}<{\centering}|p{24pt}<{\centering}||p{60pt}<{\centering}| p{56pt}<{\centering} p{46pt}<{\centering} p{46pt}<{\centering} p{46pt}<{\centering} p{46pt}<{\centering}}
 
\Xhline{1.5pt}

\multirow{2}{*}{\textbf{Dataset}} & \multirow{2}{*}{\textbf{a}}  & \multirow{2}{*}{\textbf{\baby} \textbf{(Ours)}}  & \multirow{2}{*}{\textbf{P\textsc{article}-M\textsc{ap}}}  & \multirow{2}{*}{\textbf{PML-LRS}}  & \multirow{2}{*}{\textbf{PML}-\emph{fp}} & \multirow{2}{*}{\textbf{ML-\textit{k}NN}}  & \multirow{2}{*}{\textbf{L\textsc{ift}}}  \\
  & & & & & & &\\

\Xhline{1pt}
\multicolumn{8}{c}{\multirow{1}{*}{\textbf{RLoss} $\bm{\downarrow}$}} \\

\hline
\multicolumn{1}{c|}{\multirow{4}{*}{Genbase}}
& 50	& \textbf{.005 $\pm$ .002}  	& .010 $\pm$ .002  	& .006 $\pm$ .002  	& .019 $\pm$ .006  	 & .012 $\pm$ .003  	& .009 $\pm$ .003  	\\
& 100	& \textbf{.005 $\pm$ .003}  	& .010 $\pm$ .002  	& \textbf{.005 $\pm$ .003}  	& .018 $\pm$ .007  	& .014 $\pm$ .006  	& .008 $\pm$ .004  	\\
& 150	& \textbf{.008 $\pm$ .003}  	& .010 $\pm$ .001  	& .009 $\pm$ .002  	& .019 $\pm$ .010  	 & .019 $\pm$ .006  	& .012 $\pm$ .004  	\\
& 200	& \textbf{.007 $\pm$ .003}  	& .009 $\pm$ .001  	& .010 $\pm$ .003  	& .014 $\pm$ .004  	 & .017 $\pm$ .005  	& .012 $\pm$ .005  	\\

\hline
\multicolumn{1}{c|}{\multirow{4}{*}{Medical}}
& 50	& \textbf{.028 $\pm$ .005}  	& .048 $\pm$ .005  	& .033 $\pm$ .006  	& .042 $\pm$ .009  	 & .075 $\pm$ .009  	& .044 $\pm$ .005  	\\
& 100	& \textbf{.030 $\pm$ .006}  	& .050 $\pm$ .006  	& .032 $\pm$ .006  	& .043 $\pm$ .011  	 & .078 $\pm$ .012  	& .046 $\pm$ .007  	\\
& 150	& \textbf{.034 $\pm$ .007}  	& .054 $\pm$ .005  	& .035 $\pm$ .007  	& .042 $\pm$ .009  	 & .094 $\pm$ .012  	& .055 $\pm$ .006  	\\
& 200	& \textbf{.031 $\pm$ .005}  	& .049 $\pm$ .004  	& .035 $\pm$ .007  	& .043 $\pm$ .015  	 & .088 $\pm$ .012  	& .056 $\pm$ .008  	\\

\hline
\multicolumn{1}{c|}{\multirow{4}{*}{Arts}}
& 50	& .154 $\pm$ .003  	& .142 $\pm$ .002  	& .162 $\pm$ .002  	& \textbf{.132 $\pm$ .002}  	 & .165 $\pm$ .003  	& .137 $\pm$ .003  	\\
& 100	& .162 $\pm$ .004  	& .152 $\pm$ .002  	& .170 $\pm$ .005  	& \textbf{.131 $\pm$ .002}  	 & .166 $\pm$ .003  	& .143 $\pm$ .005  	\\
& 150	& .175 $\pm$ .002  	& .158 $\pm$ .003  	& .186 $\pm$ .003  	& \textbf{.140 $\pm$ .004}  	 & .174 $\pm$ .005  	& .155 $\pm$ .003  	\\
& 200	& .180 $\pm$ .002  	& .165 $\pm$ .003  	& .192 $\pm$ .003  	& \textbf{.146 $\pm$ .001}  	 & .172 $\pm$ .004  	& .156 $\pm$ .004   \\

\hline
\multicolumn{1}{c|}{\multirow{4}{*}{Corel5k}}
& 50	& .174 $\pm$ .002  	& \textbf{.128 $\pm$ .002}  	& .206 $\pm$ .002  	& .198 $\pm$ .006  	 & .146 $\pm$ .002  	& .144 $\pm$ .001	\\
& 100	& .179 $\pm$ .003  	& \textbf{.132 $\pm$ .002}  	& .216 $\pm$ .003  	& .178 $\pm$ .008  	 & .152 $\pm$ .001  	& .154 $\pm$ .002	\\
& 150	& .185 $\pm$ .003  	& \textbf{.134 $\pm$ .002}  	& .230 $\pm$ .004  	& .169 $\pm$ .003  	 & .156 $\pm$ .002  	& .164 $\pm$ .002	\\
& 200	& .186 $\pm$ .002  	& \textbf{.135 $\pm$ .002}  	& .232 $\pm$ .003  	& .176 $\pm$ .005  	 & .161 $\pm$ .003  	& .175 $\pm$ .003	\\

\hline
\multicolumn{1}{c|}{\multirow{4}{*}{Bibtex}}
& 50	& \textbf{.094 $\pm$ .002}  	& .190 $\pm$ .004  	& .126 $\pm$ .003  	& .112 $\pm$ .004   	 & .240 $\pm$ .002 	& .121 $\pm$ .003	\\
& 100	& \textbf{.100 $\pm$ .002}  	& .187 $\pm$ .003  	& .138 $\pm$ .002  	& .107 $\pm$ .003  	 & .250 $\pm$ .003  	& .130 $\pm$ .004	\\
& 150	& .112 $\pm$ .002  	& .187 $\pm$ .002  	& .157 $\pm$ .002  	& \textbf{.109 $\pm$ .003}  	 & .260 $\pm$ .003  	& .143 $\pm$ .003	\\
& 200	& .116 $\pm$ .001  	& .189 $\pm$ .005  	& .165 $\pm$ .001  	& \textbf{.111 $\pm$ .001}  	 & .266 $\pm$ .009  	& .145 $\pm$ .002	\\

\Xhline{1pt}
\multicolumn{8}{c}{\multirow{1}{*}{\textbf{AP} $\bm{\uparrow}$}} \\

\hline
\multicolumn{1}{c|}{\multirow{4}{*}{Genbase}}
& 50	& \textbf{.991 $\pm$ .005}  	& .978 $\pm$ .005  	& .988 $\pm$ .004  	& .981 $\pm$ .006  	 & .968 $\pm$ .007  	& .982 $\pm$ .006  	\\
& 100	& \textbf{.992 $\pm$ .004}  	& .978 $\pm$ .004  	& .989 $\pm$ .004  	& .981 $\pm$ .007  	 & .967 $\pm$ .008  	& .984 $\pm$ .003  	\\
& 150	& \textbf{.988 $\pm$ .003}  	& .978 $\pm$ .004  	& .979 $\pm$ .002  	& .976 $\pm$ .011  	 & .958 $\pm$ .011  	& .979 $\pm$ .008  	\\
& 200	& \textbf{.989 $\pm$ .003}  	& .979 $\pm$ .003  	& .981 $\pm$ .005  	& .979 $\pm$ .006  	 & .964 $\pm$ .014  	& .977 $\pm$ .011  	\\

\hline
\multicolumn{1}{c|}{\multirow{4}{*}{Medical}}
& 50	& \textbf{.882 $\pm$ .014}  	& .798 $\pm$ .018  	& .853 $\pm$ .021  	& .852 $\pm$ .023  	 & .742 $\pm$ .029  	& .830 $\pm$ .017  	\\
& 100	& \textbf{.881 $\pm$ .018}  	& .791 $\pm$ .021  	& .861 $\pm$ .021  	& .855 $\pm$ .022  	 & .741 $\pm$ .029  	& .822 $\pm$ .015  	\\
& 150	& \textbf{.867 $\pm$ .022}  	& .781 $\pm$ .024  	& .858 $\pm$ .018  	& .845 $\pm$ .019  	 & .720 $\pm$ .033  	& .804 $\pm$ .006  	\\
& 200	& \textbf{.870 $\pm$ .016}  	& .768 $\pm$ .013  	& .855 $\pm$ .020  	& .853 $\pm$ .029  	 & .715 $\pm$ .034  	& .797 $\pm$ .020  	\\

\hline
\multicolumn{1}{c|}{\multirow{4}{*}{Arts}}
& 50	& \textbf{.598 $\pm$ .003}  	& .528 $\pm$ .004  	& .588 $\pm$ .003  	& .577 $\pm$ .007  	 & .488 $\pm$ .005  	& .595 $\pm$ .005  	\\
& 100	& \textbf{.597 $\pm$ .004}  	& .513 $\pm$ .004  	& .584 $\pm$ .005  	& .578 $\pm$ .005  	 & .486 $\pm$ .005  	& .591 $\pm$ .007  	\\
& 150	& \textbf{.577 $\pm$ .005}  	& .499 $\pm$ .006  	& .564 $\pm$ .005  	& .558 $\pm$ .005  	 & .478 $\pm$ .006  	& \textbf{.577 $\pm$ .003}  	\\
& 200	& .572 $\pm$ .003  	& .491 $\pm$ .006  	& .557 $\pm$ .005  	& .554 $\pm$ .005  	& .477 $\pm$ .003  	& \textbf{.578 $\pm$ .005}  	\\

\hline
\multicolumn{1}{c|}{\multirow{4}{*}{Corel5k}}
& 50	& \textbf{.295 $\pm$ .004}  	& .263 $\pm$ .005  	& .282 $\pm$ .003  	& .240 $\pm$ .003  	 & .233 $\pm$ .003  	& .244 $\pm$ .005	\\
& 100	& \textbf{.293 $\pm$ .004}  	& .260 $\pm$ .003  	& .276 $\pm$ .004  	& .242 $\pm$ .003  	 & .229 $\pm$ .003  	& .217 $\pm$ .004	\\
& 150	& \textbf{.289 $\pm$ .004}  	& .264 $\pm$ .005  	& .266 $\pm$ .003  	& .241 $\pm$ .003   	 & .226 $\pm$ .003 	& .194 $\pm$ .003	\\
& 200	& \textbf{.288 $\pm$ .004}  	& .260 $\pm$ .004  	& .266 $\pm$ .004  	& .241 $\pm$ .003  	 & .224 $\pm$ .003  	& .185 $\pm$ .005	\\

\hline
\multicolumn{1}{c|}{\multirow{4}{*}{Bibtex}}
& 50	& \textbf{.567 $\pm$ .004}  	& .383 $\pm$ .007  	& .532 $\pm$ .003  	& .517 $\pm$ .003  	 & .295 $\pm$ .004  	& .487 $\pm$ .007	\\
& 100	& \textbf{.555 $\pm$ .004}  	& .380 $\pm$ .006  	& .509 $\pm$ .005  	& .522 $\pm$ .005  	 & .282 $\pm$ .005  	& .467 $\pm$ .007	\\
& 150	& \textbf{.536 $\pm$ .004}  	& .369 $\pm$ .003  	& .476 $\pm$ .006  	& .519 $\pm$ .004  	 & .270 $\pm$ .004  	& .448 $\pm$ .007	\\
& 200	& \textbf{.528 $\pm$ .006}  	& .365 $\pm$ .004  	& .460 $\pm$ .005  	& .511 $\pm$ .006  	 & .266 $\pm$ .005  	& .440 $\pm$ .007	\\

\Xhline{1pt}
\multicolumn{8}{c}{\multirow{1}{*}{\textbf{Macro-F1} $\bm{\uparrow}$}} \\

\hline
\multicolumn{1}{c|}{\multirow{4}{*}{Genbase}}
& 50	& \textbf{.710 $\pm$ .029}  	& .543 $\pm$ .053  	& .680 $\pm$ .015  	& .598 $\pm$ .023  	 & .622 $\pm$ .022  	& .619 $\pm$ .044  	\\
& 100	& \textbf{.722 $\pm$ .044}  	& .522 $\pm$ .017  	& .710 $\pm$ .027  	& .594 $\pm$ .056  	 & .594 $\pm$ .022  	& .600 $\pm$ .039  	\\
& 150	& \textbf{.649 $\pm$ .033}  	& .536 $\pm$ .034  	& .618 $\pm$ .039  	& .603 $\pm$ .030  	 & .540 $\pm$ .016  	& .566 $\pm$ .046  	\\
& 200	& \textbf{.652 $\pm$ .054}  	& .533 $\pm$ .019  	& .559 $\pm$ .033  	& .603 $\pm$ .026  	 & .562 $\pm$ .058  	& .579 $\pm$ .039  	\\

\hline
\multicolumn{1}{c|}{\multirow{4}{*}{Medical}}
& 50	& \textbf{.405 $\pm$ .027}  	& .270 $\pm$ .013  	& .301 $\pm$ .016  	& .296 $\pm$ .015  	 & .243 $\pm$ .034  	& .309 $\pm$ .022  	\\
& 100	& \textbf{.363 $\pm$ .017}  	& .254 $\pm$ .015  	& .314 $\pm$ .021  	& .320 $\pm$ .020  	 & .235 $\pm$ .020  	& .293 $\pm$ .015  	\\
& 150	& \textbf{.348 $\pm$ .026}  	& .238 $\pm$ .006  	& .316 $\pm$ .013  	& .294 $\pm$ .014  	 & .208 $\pm$ .016  	& .264 $\pm$ .010  	\\
& 200	& \textbf{.373 $\pm$ .024}  	& .227 $\pm$ .016  	& .315 $\pm$ .015  	& .315 $\pm$ .036  	 & .192 $\pm$ .019  	& .266 $\pm$ .020  	\\

\hline
\multicolumn{1}{c|}{\multirow{4}{*}{Arts}}
& 50	& .244 $\pm$ .012  	& .201 $\pm$ .007  	& .220 $\pm$ .008  	& .159 $\pm$ .008  	& .123 $\pm$ .007  	& \textbf{.249 $\pm$ .009}  	\\
& 100	& \textbf{.251 $\pm$ .012}  	& .190 $\pm$ .004  	& .227 $\pm$ .006  	& .156 $\pm$ .004  	 & .119 $\pm$ .008  	& .247 $\pm$ .006  	\\
& 150	& \textbf{.240 $\pm$ .007}  	& .180 $\pm$ .005  	& .229 $\pm$ .006  	& .129 $\pm$ .004  	 & .112 $\pm$ .010  	& .237 $\pm$ .009  	\\
& 200	& \textbf{.226 $\pm$ .006}  	& .184 $\pm$ .007  	& .210 $\pm$ .006  	& .126 $\pm$ .004  	 & .108 $\pm$ .005  	& .217 $\pm$ .005  	\\

\hline
\multicolumn{1}{c|}{\multirow{4}{*}{Corel5k}}
& 50	& .040 $\pm$ .001  	& .027 $\pm$ .001  	& .039 $\pm$ .001  	& .005 $\pm$ .000  	& .020 $\pm$ .001  	& \textbf{.046 $\pm$ .002}	\\
& 100	& .038 $\pm$ .000  	& .028 $\pm$ .002  	& .038 $\pm$ .001  	& .004 $\pm$ .000  	& .020 $\pm$ .002  	& \textbf{.040 $\pm$ .002}	\\
& 150	& \textbf{.037 $\pm$ .001}  	& .034 $\pm$ .002  	& \textbf{.037 $\pm$ .001}  	& .004 $\pm$ .000  	& .019 $\pm$ .002  	& .035 $\pm$ .002	\\
& 200	& \textbf{.036 $\pm$ .000}  	& .032 $\pm$ .002  	& .032 $\pm$ .000  	& .004 $\pm$ .000  	 & .018 $\pm$ .001  	& .033 $\pm$ .001	\\

\hline
\multicolumn{1}{c|}{\multirow{4}{*}{Bibtex}}
& 50	& \textbf{.375 $\pm$ .002}  	& .163 $\pm$ .004  	& .359 $\pm$ .004  	& .299 $\pm$ .006  	 & .133 $\pm$ .001  	& .299 $\pm$ .008	\\
& 100	& \textbf{.360 $\pm$ .002}  	& .157 $\pm$ .004  	& .332 $\pm$ .004  	& .311 $\pm$ .002  	 & .120 $\pm$ .004  	& .271 $\pm$ .008\\
& 150	& \textbf{.340 $\pm$ .002}  	& .146 $\pm$ .003  	& .296 $\pm$ .006  	& .306 $\pm$ .007  	 & .109 $\pm$ .001  	& .247 $\pm$ .009	\\
& 200	& \textbf{.331 $\pm$ .004}  	& .145 $\pm$ .005  	& .278 $\pm$ .004  	& .301 $\pm$ .002  	 & .103 $\pm$ .005  	& .232 $\pm$ .010	\\

\Xhline{1.5pt}
\end{tabular}
\caption{Experimental results (mean $\pm$ std) in terms of \textit{RLoss}, \textit{AP}, \textit{Macro-F1}, where the best performance is shown in boldface.}
\label{R1}
\end{table*}

\subsubsection{Datasets}
In the experiment, we downloaded five public multi-label datasets from the \emph{mulan} website\footnote{http://mulan.sourceforge.net/datasets-mlc.html}, including \emph{Genbase}, \emph{Medical}, \emph{Arts}, \emph{Corel5k} and \emph{Bibtex}. The statistics of those datasets are outlined in Table \ref{Datasets}.

To conduct experiments under the scenario of noisy supervision, we generate synthetic PML datasets from each original dataset by randomly drawing irrelevant noisy labels. Specifically, for each instance, we create the candidate label set by adding several randomly drawn irrelevant labels with $a\%$ number of ground-truth labels, where we vary $a$ over the range $\{50,100,150,200\}$. Besides, to avoid useless PML instances annotated with all $l$ labels, we fix each candidate label set at most $l-1$ labels. Accordingly, a total of twenty synthetic PML datasets are generated.

\subsubsection{Baseline Methods}
We employed five baseline methods for comparison, including three PML methods and two traditional methods of Multi-label Learning (ML). For the ML ones, we directly train them over the synthetic PML datasets by considering all candidate labels as ground-truth ones. We outline the method-specific settings below.

\begin{itemize}
\item \textbf{PML-\emph{fp}} \cite{PML2018}: A PML method with a ranking loss objective weighted by ground-truth confidences. We utilize the code provided by its authors, and tune the parameters following the original paper. Here, the other version of \cite{PML2018}, \ie PML-\emph{lc}, was neglected, since it performed worse than PML-\emph{fp} in our early experiments.
\item \textbf{P\textsc{article}-M\textsc{ap}} \cite{PML2019}: A two-stage PML method with label propagation. We employ the public code\footnote{http://palm.seu.edu.cn/zhangml/files/PARTICLE.rar}, and tune the parameters following the original paper. Here, the other version of \cite{PML2019}, \ie P\textsc{article}-V\textsc{ls}, was neglected, since it performed worse than P\textsc{article}-M\textsc{ap} in our early experiments.
\item \textbf{PML-LRS} \cite{Juan2019}: A PML method with candidate label decomposition. We utilize the code provided by its authors, and tune the regularization parameters over $\{10^i|i=-3,\cdots,1\}$ using cross-validations.
\item \textbf{M}ulti-\textbf{L}abel \textbf{\emph{k}} \textbf{N}earest \textbf{N}eighbor (\textbf{ML-\textit{k}NN}) \cite{MLKNN2007}: A \emph{k}NN ML method. We employ the public code\footnote{http://palm.seu.edu.cn/zhangml/files/ML-kNN.rar} implemented by its authors, and tune its parameters following the original paper.
\item Multi-label learning with \textbf{L}abel spec\textbf{I}fic \textbf{F}ea\textbf{T}ures
 (\textbf{L\textsc{ift})} \cite{LIFT2015}: A binary relevance ML method. We employ the public code\footnote{http://palm.seu.edu.cn/zhangml/files/LIFT.rar} implemented by its authors, and tune its parameters following the original paper.
\end{itemize}

For our \baby, the regularization parameter $\lambda_1$ is fixed to 1, and $\lambda_2$ is tuned over $\{10^i|i=1,2\}$ using 5-fold cross validation results.

\subsubsection{Evaluation Metrics}
We employed seven evaluation metrics \cite{TKDE2014}, including \emph{Subset Accuracy} (SAccuracy), \emph{Hamming Loss} (HLoss), \emph{One Error} (OError), \emph{Ranking Loss} (RLoss), \emph{Average Precision} (AP), \emph{Macro Averaging F1} (Macro-F1), \emph{Micro Averaging F1} (Micro-F1), where both instance-based and label-based metrics are covered. For SAccuracy, AP, Macro-F1 and Micro-F1, the higher value is better, while for HLoss, OError and RLoss, a smaller value is better.

\begin{table*}[ht]
\small
  \renewcommand\arraystretch{1.2}
  \label{table5-4}
  \begin{center}
      {
        \begin{tabular}{c||ccccccc|c}
        \Xhline{1.5pt}
        {Baseline Method} &{SAccuracy} &{HLoss} &{OError} &{RLoss} &{AP} &{Macro-F1} &{Micro-F1} &\textbf{Total} \\
        \hline\hline
        {P\textsc{article}}-{M\textsc{ap}} &{20/0/0} &{16/4/0} &{20/0/0} &{12/0/8} &{20/0/0} &{20/0/0} &{20/0/0} &\textbf{128/4/8} \\
        {PML-LRS} &{17/2/1} &{9/11/0} &{20/0/0} &{17/3/0} &{20/0/0} &{18/2/0} &{20/0/0} &\textbf{121/18/1} \\
        {PML}-\emph{fp} &{20/0/0} &{18/2/0} &{19/1/0} &{11/1/8} &{20/0/0} &{20/0/0} &{20/0/0} &\textbf{128/4/8} \\
        {ML-\textit{k}NN} &{20/0/0} &{20/0/0} &{20/0/0} &{14/0/6} &{20/0/0} &{20/0/0} &{20/0/0} &\textbf{134/0/6} \\
        {L\textsc{ift}} &{19/0/1} &{17/3/0} &{17/2/1} &{12/0/8} &{18/1/1} &{17/0/3} &{19/1/0} &\textbf{119/7/14} \\
        \Xhline{1.5pt}
      \end{tabular}
     }
\end{center}
\caption{Win/tie/loss counts of pairwise $t$-test (at 0.05 significance level) between \baby and each comparing approach.}
\label{R2}
\end{table*}

\subsection{Experimental Results}
\label{4.2}

For each PML dataset, we randomly generate five 50\%/50\% training/test splits, and evaluate the average scores ($\pm$standard deviation) of comparing algorithms. Due to the space limitation, we only present detailed results of RLoss, AP and Macro-F1 in Table \ref{R1}, while the observations of other metrics tend to be similar. First, we can observe that \baby outperforms other three PML methods in most cases, where \baby dominates the scenarios of AP and Macro-F1 on different noise levels. Especially, the performance gain over P\textsc{article}-M\textsc{ap} indicates that using the information from non-candidate labels is beneficial for PML. Besides, we can see that \baby significantly performs better than the two traditional ML methods in most cases, since they directly use noisy candidate labels for training.        

Additionally, for each PML dataset and evaluation metric, we conduct a pairwise \emph{t}-test (at 0.05 significance level) to examine whether \baby is statistically different to baselines. The win/tie/loss counts over 20 PML datasets and 7 evaluation metrics are presented in Table \ref{R2}. We can observe that \baby significantly outperforms the PML baseline methods P\textsc{article}-M\textsc{ap}, PML-LRS and PML-\emph{fp} in $91.4\%$, $86.4\%$ and $91.4\%$ cases, and also outperforms the two traditional MLL methods ML-\emph{k}NN and L\textsc{ift} in $95.7\%$ and $85\%$ cases. Besides, on the results of evaluation metrics, \baby achieves significantly better scores. For example, the SAccuracy, OError, AP, Macro-F1 and Micro-F1 of \baby are better than all comparing algorithms in $95\%$ cases.

\begin{figure}[t]
\includegraphics[width=0.45\textwidth]{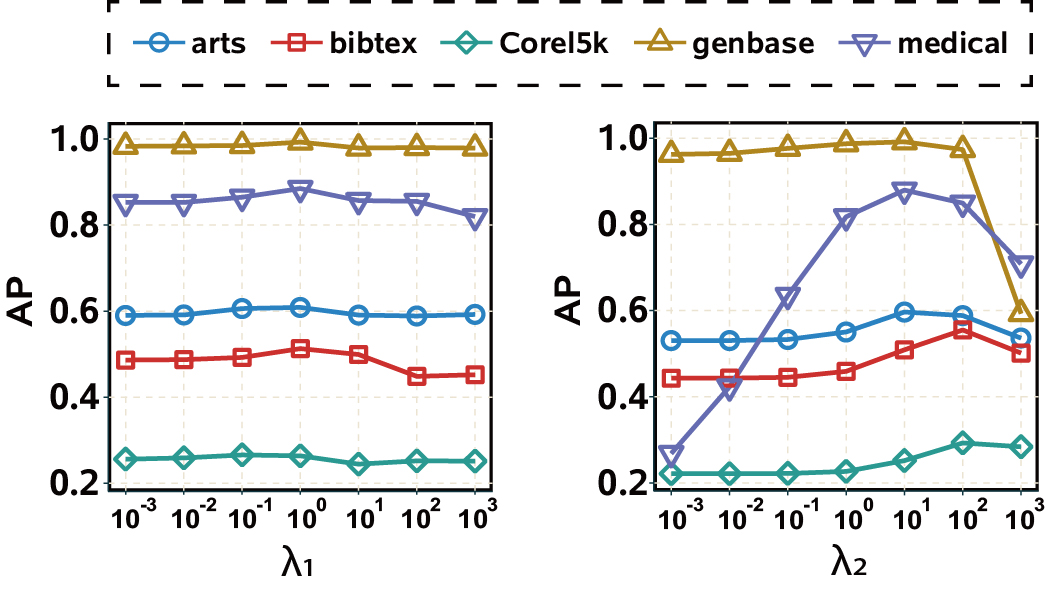}
\centering
\caption{Sensitivity analysis of the regularization parameters $\{\lambda_1,\lambda_2\}$}
\label{FigLambda}
\end{figure}

\subsection{Parameter Sensitivity}
\label{4.3}

We empirically analyze the sensitivities of the regularization parameters $\{\lambda_1,\lambda_2\}$ of \baby. For each one, we examine the AP scores by varying its value from $\{10^i|i=-3,\cdots,3\}$ across PML datasets with $a=100$ by holding the other two fixed. The experimental results are presented in Fig.\ref{FigLambda}. First, we can observe that the AP scores of $\lambda_1$ seem quite stable over different types of PML datasets. That is, we empirically conclude that \baby is insensitive to $\lambda_1$, making \baby more practical. Second, for $\lambda_2$, it performs better with the values of $\{10^i|i=1,2\}$ across all PML datasets, which are the settings used in our experiment.

\section{Conclusion}
\label{5}

We concentrate on the task of PML, and propose a novel two-stage \baby algorithm. In the first stage, \baby performs an unconstrained label propagation procedure to estimate the label enrichment, which simultaneously involves the relevance degrees of candidate labels and irrelevance degrees of non-candidate labels. In the second stage, \baby jointly learns the ground-truth confidence and multi-label predictor
given the label enrichment. Extensive experiments on PML datasets indicate the superior performance of \baby.

\section*{Acknowledgments}

We would like to acknowledge support for this project from the National Natural Science Foundation of China (NSFC) (No.61602204, No.61876071, No.61806035, No.U1936217)

\bibliographystyle{named}
\bibliography{PML}

\end{document}